%% file: main.tex
\documentclass[sigconf]{acmart}
\usepackage{latexsym}
\usepackage{microtype}
\usepackage{graphicx}
\usepackage{subfigure}
\usepackage{booktabs} 
\usepackage{amsmath}
\usepackage{tabularx}
\usepackage{dsfont}
\usepackage{float}
\usepackage{placeins}
\usepackage{afterpage}
\usepackage{multirow}
\usepackage{multicol}
\usepackage[hide]{notes-alt}
\usepackage{tikz}

\usepackage{microtype}

\author{Zijian Zhang}
\affiliation{%
   \institution{L3S Research Center}
   \city{Hannover}
   \country{Germany}
 } 
\email{zzhang@l3s.de}

\author{Koustav Rudra}
\affiliation{%
   \institution{L3S Research Center}
   \city{Hannover}
   \country{Germany}
 } 
\email{rudra@l3s.de}

\author{Avishek Anand}
\affiliation{%
   \institution{L3S Research Center}
   \city{Hannover}
   \country{Germany}
 } 
\email{anand@l3s.de}

\date{}

\newcommand{\movie}{Movie Reviews}
\newcommand{\multirc}{MultiRC}
\newcommand{\fever}{FEVER}

\newcommand{\bert}{\textsc{Bert}}
\newcommand{\bertbase}{\textsc{Bert}_\textrm{Base}}

\newcommand{\approach}{\textsc{ExPred}}
\newcommand{\methodindv}{\textsc{ExPred-Soft}}

\newcommand{\mtlsoft}{\textsc{ExPred-Soft}}
\newcommand{\mtlhard}{\textsc{ExPred}}

\newcommand{\exptask}{\textsc{Expred (w/o Task Sup.)}}
\newcommand{\mtltask}{\textsc{Expred-Stage-1}}

\newcommand{\human}{\textsc{Human Explanation}}
\newcommand{\cls}{\textsc{Full Input}}

\newcommand{\mpara}[1]{\medskip\noindent{\bf #1}}
\copyrightyear{2021}
\acmYear{2021}
\setcopyright{acmcopyright}\acmConference[WSDM '21]{Proceedings of the Fourteenth ACM International Conference on Web Search and Data Mining}{March 8--12, 2021}{Virtual Event, Israel}
\acmBooktitle{Proceedings of the Fourteenth ACM International Conference on Web Search and Data Mining (WSDM '21), March 8--12, 2021, Virtual Event, Israel}
\acmPrice{15.00}
\acmDOI{10.1145/3437963.3441758}
\acmISBN{978-1-4503-8297-7/21/03}
\begin{document}
\fancyhead{}

\title{Explain and Predict, and then Predict Again}

\input{00.abstract.tex}


\maketitle

\input{intro-avi-v2}

\input{relwork-avi-v2}

\input{04.approach.tex}

\input{setup-avi}

\input{results-avi}

\input{analysis-avi}

\input{07.conclusions.tex}

~\\
\noindent \textbf{Acknowledgement:} Funding for this project was in part provided by the European Union’s Horizon 2020 research and innovation program under grant agreement No 832921, and No 871042.

\bibliography{emnlp2020}
\bibliographystyle{acl_natbib}

\end{document}

%% file: 00.abstract.tex
\begin{abstract}
    
A desirable property of learning systems is to be both effective and interpretable. 
Towards this goal, recent models have been proposed that first generate an extractive explanation from the input text and then generate a prediction on just the explanation called \textit{explain-then-predict models}.
These models primarily consider the task input as a supervision signal in learning an extractive explanation and do not effectively integrate \textit{rationales data} as an additional inductive bias to improve task performance.

We propose a novel yet simple approach \approach{}, which uses multi-task learning in the explanation generation phase effectively trading-off explanation and prediction losses. Next, we use another prediction network on just the extracted explanations for optimizing the task performance. We conduct an extensive evaluation of our approach on three diverse language datasets -- sentiment classification, fact-checking, and question answering -- and find that we substantially outperform existing approaches.

\end{abstract}

%% file: intro-avi-v2.tex
\section{Introduction}
\label{sec:intro}

Web content analysis using text has been recently dominated by complex representation learning approaches using neural models. 
A key concern using complex learning systems is regarding their interpretability in that it is hard to determine if the predictions of these models are grounded in the right reasons.
Towards this, there has been an upsurge of approaches that intend to interpret the decisions of complex learning models using post-hoc analysis~\cite{ribeiro2016should:lime,lundberg2017unified,singh2020model}. A key problem in post-hoc interpretability is in its inherent uncertainty of the evaluation, that is -- ground truth for the actual machine rationale behind a certain decision is missing. The alternate design philosophy is to construct models that are \textit{interpretable by design}, obviating the need for post-hoc interpretability, that produce an explanation or rationale along with the decision~\cite{lei2016rationalizing,lehman2019inferring}. 

\begin{figure}[t]
    \centering
    \includegraphics[width=0.8\linewidth]{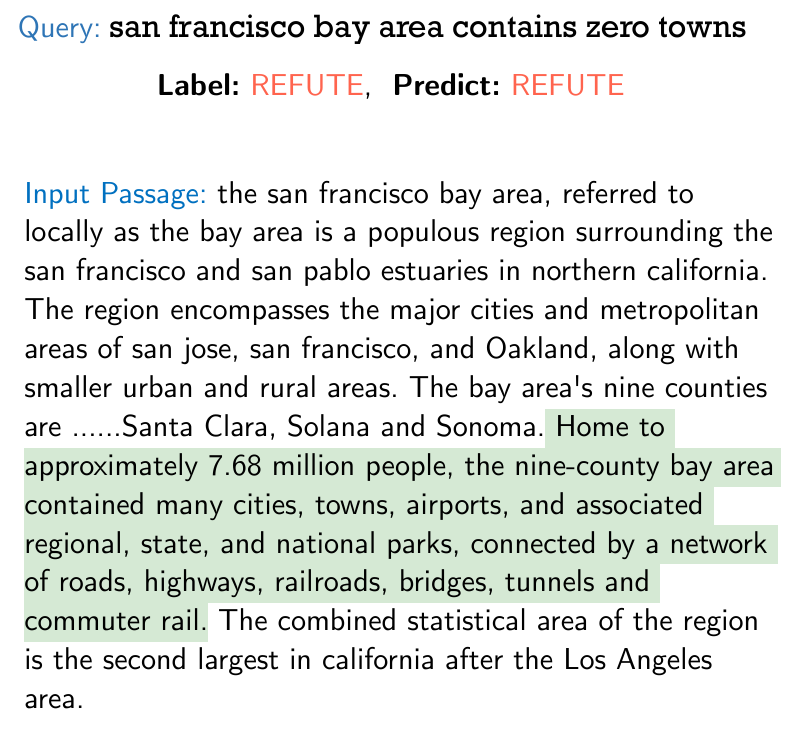}
    \caption{\small{ An anecdotal example of an extractive explanation of our \approach{} model that refutes the query using a passage from the \fever{} dataset. The explanation is highlighted in green.}}
    \label{fig:anecdotal_example}
\end{figure}

This paper aims to learn accurate models that are \emph{interpretable by design} by effectively using ``rationales'' data.
A rationale is defined to be a small yet sufficient part of the input text, short so that it makes clear what is most important, and sufficient so that a correct prediction can be made from the rationale alone~\cite{bastings2019interpretable}.
For many language tasks found in the Web, like fact-checking, sentiment detection, and question answering, \emph{rationales} are available that encode human reasoning in the form of extractive task-specific summaries, as shown in Figure~\ref{fig:anecdotal_example}.
Rationale data has been used to improve the performance of prediction tasks~\cite{zaidan2008modeling,zhang2016rationale,ross:17:right}, but these models do not generate explanations.

{}

We are specifically interested in models where explanations are first-class citizens, in that each prediction can be unambiguously attributed to a reason or rationale that is human-understandable.
Towards this, we focus on a recently proposed framework that we refer to as \emph{explain-then-predict} models~\cite{lei2016rationalizing,bastings2019interpretable,lehman2019inferring} . Such models perform the task prediction in a two-stage manner. In the explanation phase, a model learns to extract the rationale from the input text. 
In the subsequent prediction phase, another independent model predicts the task output solely based on the extractive explanation. 
Unlike the post-hoc approaches, the explain-then-predict setup unambiguously attributes the reason for a given prediction to the extractive explanations. 

A crucial limitation of explain-then-predict models is that they fail to learn accurate models since they either ignore or do not effectively utilize the rationales data as a supervision signal.
Specifically, ~\citet{lei2016rationalizing,bastings2019interpretable,yoon2018:invase} train end-to-end models that are only supervised on task-specific training data. 
On the other hand,~\citet{lehman2019inferring} follows a pipelined approach that explicitly uses the rationales data in the explanation generation phase but is agnostic to task-specific signals, thus not being able to generalize well in the subsequent prediction phase.
This paper's main objective is to exploit supervision signals from both rationales data and task objective for generating \emph{task-aware} explanations to improve task performance.

Unlike earlier approaches, our idea is simple -- we learn to generate explanations supervised by both task-specific and rationale-based signals in our explanation generation phase.
We realize this by using multi-task learning, where \emph{task prediction} and \emph{explanation generation} are both learned on a common encoder substrate (cf. Figure 2). 
After training the explanation model, in the prediction phase, a separately parameterized model for task prediction is learned just on the generated explanation. 
We refer to this scheme of predicting and explaining first (in the explanation generation phase) and then predicting again (in the prediction phase) as \approach{}.

We conduct an extensive evaluation of \approach{} on three different language tasks found in the Web, where human rationales are provided -- \emph{sentiment classification, fact checking} and \emph{question answering}.
We find that using a shared representation space for encoding the input for prediction and explanation generation results in more task-specific explanations.  
We also observe that \approach{} can effectively balance the task and explanation performance by learning to generate task-specific explanations.

\mpara{Our contributions.} In sum the key contributions of our work are 

\begin{itemize}
    \item 
    We propose a novel explanation generation framework work using multi-task learning~\approach{} that is task-aware and can exploit rationales data for effective explanations.
    
    \item 
    We show that our explanations show significant improvements in task performance (up to $7\%$) and explanation accuracy (up to $20\%$) over existing baselines.
\end{itemize}

For the sake of reproducibility, the code for the experiments described in this paper will be made available at \url{https://github.com/JoshuaGhost/expred}.

%% file: relwork-avi-v2.tex
\section{Related Work}
\label{sec:rel-work}

Classical models are known to exhibit a natural trade-off between task performance and being interpretable. As a result, in recently popular post-hoc interpretability approaches that do not negotiate task performance and instead rely on interpreting already trained models in a post-hoc manner~\cite{lundberg2017unified,ribeiro2016should:lime,xu2015show,koh2017understanding}.
However, a fundamental limitation of such post-hoc approaches is that -- explanations might be faithful to the predictions of the model but might not be faithful to the model's actual decision-making process of the model~\cite{rudin2019stop} or to human reasoning~\cite{dissonance19:anand}. Secondly, and more worrisome is the problem of evaluation of interpretability techniques due to difficulty in gathering ground truth for evaluating an explanation due to human bias~\cite{lage2019evaluation}. 

\mpara{Explain-then-predict models.} \citet{lei2016rationalizing} proposed a sequential approach of rationale generation followed by prediction using the generated prediction. Similar frameworks that mainly differ in how they perform end-to-end training due to the explanation sampling step have been proposed subsequently. 
Common proposals for training include using REINFORCE~\cite{lei2016rationalizing}, actor-critic methods~\cite{yoon2018:invase}, or re-parameterization tricks~\cite{bastings2019interpretable}.
\citet{lehman2019inferring} uses a similar philosophy of decoupling rationale generator and predictor, albeit a slightly different architecture and supervised using human rationales. Instead, we explicitly use human rationales to provide the supervision signal and decouple the prediction network from the explanation phase.

\mpara{Rationale-based prediction.} Related to our work is the work on rationale classification and has roots in the seminal work of~\citet{zaidan2008modeling,zaidan2007using} that aims to improve model generalization by utilizing human rationales as inductive bias.
The closest to our work of building explain-then-predict models using rationale data is~\citet{deyoung2019eraser}, who instead use rationale predictions as regularizers to the task loss. 
We use these approaches as competitors in our experiments. 

Unlike us, all these approaches are agnostic to task supervision when learning to generate explanations. 
An exception is~\citet{zhong2019fine} that showed supervising regularizes the attention layer with human annotations while learning from task supervision. 
However, it is not an explain-then-predict model. Specifically, it is hard to unambiguously attribute the rationale of the prediction since the prediction phase still has access to the input.

\mpara{Interpretability for Language Tasks.}
With the tremendous growth of the Web~\cite{holzmann2017exploring, holzmann2016dawn}, many language tasks on the Web are being treated learning tasks.
For language tasks, there has been work on post-hoc analysis of already learned neural models by analyzing state activation~\cite{NIPS2013_5166, karpathy2015visualizing,li2016understanding} or attention weights~\cite{cheng2016long, martins2016softmax, cui2016attention, yang2016hierarchical}.
The attention weights learned as weights assigned to token representation are intended to describe rationales.
However, recently the faithfulness of interpreting model prediction with soft attention weights has been called into question \cite{wiegreffe2019attention, jain2019attention}. 
Specifically, the contextual entanglement of inputs is non-trivial. The prediction model can still perform well even if the attention weights don't correlate with the (sub-)token weight as desired by humans.
Our approach for rationale based explanations differs in the type of architectures, objectives, and general nature of its utility.

%% file: 04.approach.tex
    \begin{figure}[tb]
    \centering
    \includegraphics[width=0.9\columnwidth]{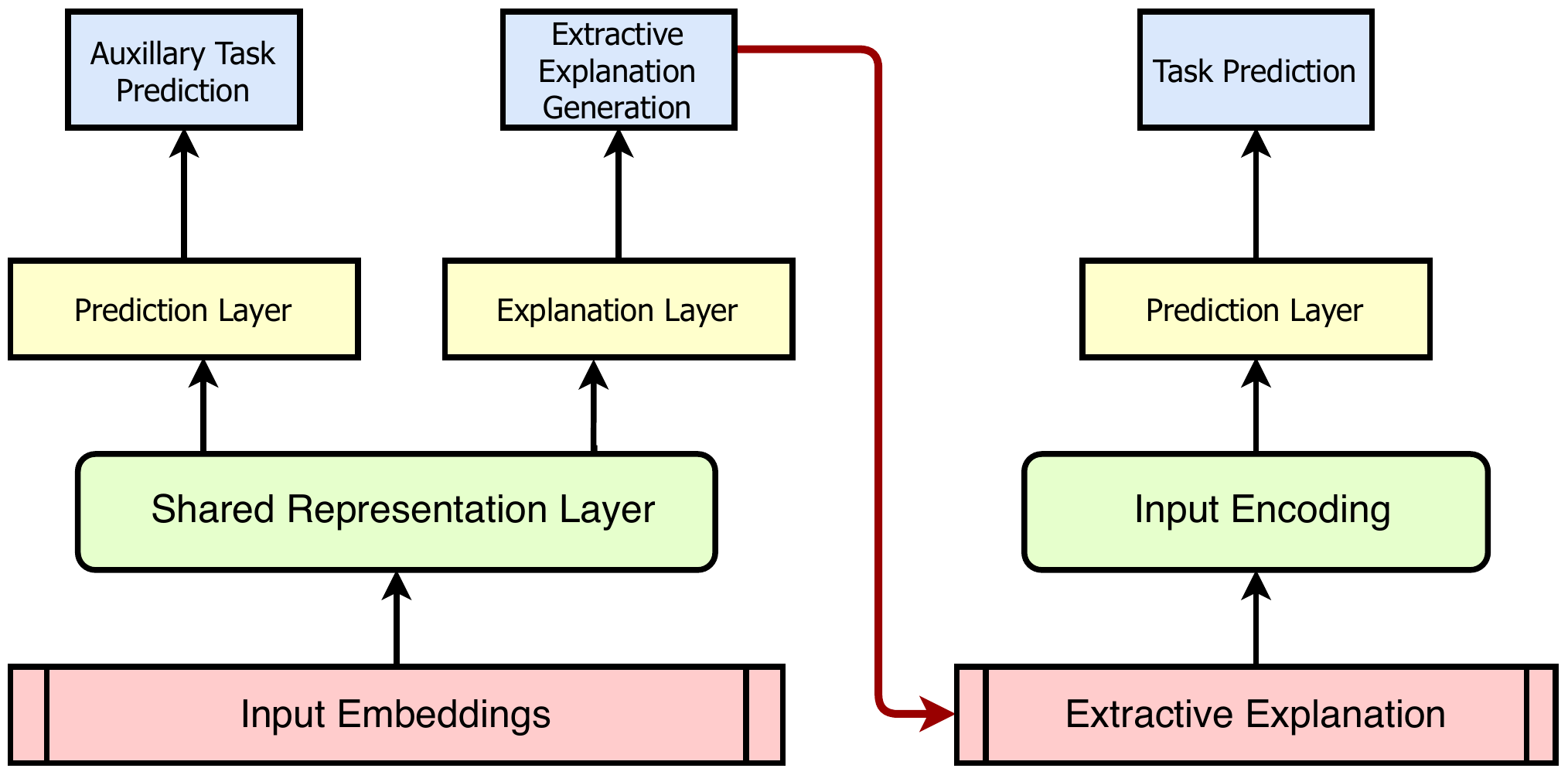}
    \caption{\small{Overview of \approach{}.  Explanation generation supervised by \textit{task objectives} and \textit{explanation generation}. Here auxiliary task is the same as the actual task (the Task Prediction on the right). }}\label{fig:mtl-example}
    \end{figure}

\section{Approach}

We aim to come up with a model that can generate explanations as well as high-quality predictions, given access to human rationales accompanying task-specific training instances.
Human rationales are sets of sequences of the input text that have been annotated by humans as potential reasons for the prediction. 

We formalize here the task of \emph{extractive rationale generation} in the context of neural models where we are provided with a sequence of words as input, namely $\mathbf{x}=\left<x^{1}, \cdots, x^{|S|}\right>$, where $|S|$ is the length of the sequence and each $x^{i} \in \mathbb{R}^{d}$ denotes the vector representation of the $i$-th word and task labels $\mathbf{y}$. 
Additionally, we also assume that each word $x^i$ has an associated Boolean label $t^i \in\{0,1\}$, where $t^i = 1$ if word $i$ is a part of the rationale else $t^i = 0$. The rationales of the sequence is then $\mathbf{t}\in\{0,1\}^{|S|}$
Typically, rationales are sequences of words and hence a potential rationale is a sub-sequence of the input sequence. Note that multiple non-overlapping sub-sequences might exist for a given input text.

\subsection{Approach Overview}
\label{sec:approach}

Our goal is to construct a explain-then-predict model that is composed of a \textbf{explanation generation network} $g^{\phi}$ parameterized by ${\phi}$ and a \textbf{prediction network} $f^{\theta}$ parameterized by ${\theta}$.
The explanation generation network $g^{\phi}$ first maps the input $\mathbf{x}$ into an explanation mask $\mathbf{t}$. Thereafter, the prediction network $f^{\theta}$ maps the masked input $\mathbf{x} \otimes \mathbf{t}$ to the task output $\mathbf{y}$.

Our key insight is that in generating effective task-specific explanations, we would ideally want to be influenced by task-specific supervision along with rationale-specific supervision.
Towards this, we propose to use the Multi-task Learning (MTL) framework~\cite{caruana1997multitask} for the explanation generation phase. 
In MTL, the original prediction task is trained along with multiple related auxiliary tasks using shared or tied parameters~\cite{liu2017entity} as a form of inductive transfer that causes a model to prefer some hypothesis over others.
This is indeed the case in our problem where the prediction and rationale generation tasks are closely related and we intend to generate a task-specific explanation.

Consequently, we introduce an auxiliary task in the explanation generation phase modeled by a \textbf{auxiliary task predictor network} $f^{\psi}$ parameterized by ${\psi}$ such that $f^{\psi}$ also maps the input $\mathbf{x}$ to task output $\mathbf{y}$.
We use the shared encoder architecture of MTL, that is, we enforce that the explanation generator $g^{\phi}$ and auxiliary task predictor $f^{\psi}$ share the same encoder $\mathbf{enc}(.)$ but different decoders
.
Note that the auxiliary task in our case is indeed the actual prediction task.

We can now conceive different models for the explanation generator $g^{\phi}$, auxiliary task predictor $f^{\psi}$, and task predictor $f^{\theta}$. 
The high-level architecture of our approach \approach{} is presented in Figure~\ref{fig:mtl-example} where we follow a pipelined architecture of explanation generation followed by the actual prediction task. 
In what follows we describe our design choices and training details for each of these networks.




\subsection{Explanation Generation}
\label{sec:explanation_generation}

In our explanation generation phase, we detail our architectural design choices for encoders and decoders and our loss function.

\subsubsection{Shared Encoder.}
Since contextualized models like BERT \cite{devlin2018bert} are now de-facto models like  for representing text input, we use the BERT model as our shared encoder $\mathbf{enc}(.)$ between $g^{\phi}$ and auxiliary task predictor $f^{\psi}$.
In principle, BERT can be replaced by any text encoding model as an encoder -- LSTMs, other transformer-based encoders, etc. 
We follow the standard practices in handling text input in \bert{}.
Specifically, a single sentence or sentence pair is fed to \bert{} based on the type of tasks.
Sentence tokens, segments, and positional information are taken as inputs. 
Technically, for a single sentence task, this is realized by forming an input to BERT of the form $[[CLS],<sentence>]$ and padding each sequence in a mini-batch to the maximum length (typically 512 tokens) in the batch. Similarly, a sentence-pair task is realized by $[[CLS],<sentence1>,[SEP],<sentence2>]$ and the entire sequence is of maximum length 512 \cite{devlin2018bert}. The final hidden state corresponding to the $[CLS]$ token captures the high-level representation of the entire text and other vectors represent the corresponding embeddings of the input tokens. Hence, we obtain a $512\times768$ dimensional representation of the input sequence where 512 is the maximum number of input tokens.

The working principle of recent auto-regressive language models is significantly better than word-based representations (word2vec) and long dependency modeling networks (RNN and LSTM)\cite{devlin2018bert}. Word2vec models assume independence between words present in a sentence that does not hold. Contextual auto-regressive neural models such as \bert~overcome that limitation. The model also works as a knowledge-base due to its pre-training over a large amount of unlabelled corpus~\cite{petroni2019language}. On the other hand, LSTM based models were proposed to capture long term dependencies among words and overcome the problem of vanishing gradient. However, this scheme does not work for large paragraphs. \bert~completely relies on self-attention instead of multiple gates. This increases the complexity quadratically but helps to capture the interaction between each pair of words. 

\subsubsection{Decoders}
\begin{figure}[t]
    \centering
    \includegraphics[width=0.8\columnwidth]{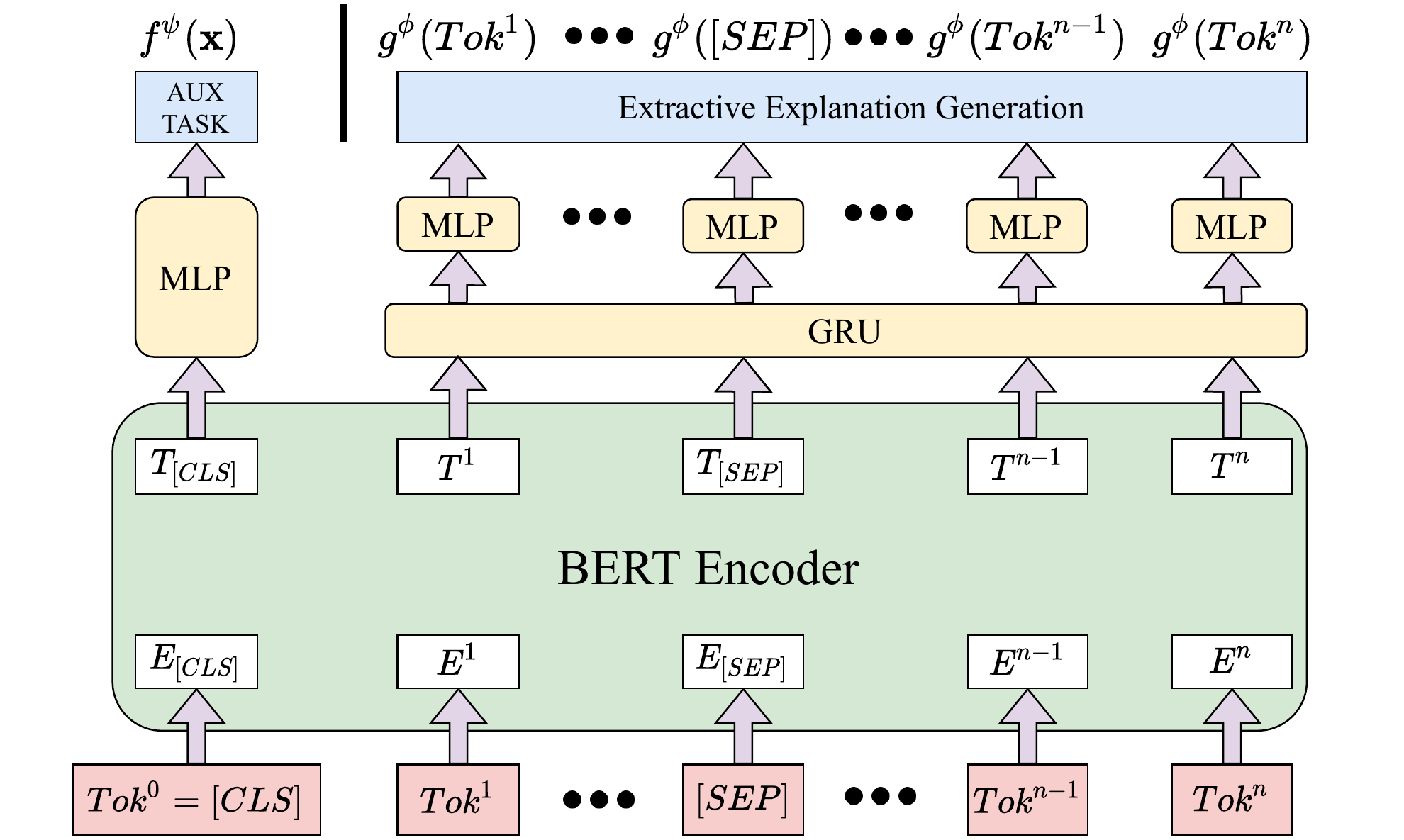}
    
    \caption{ Multi-task learning for joint optimization of \textit{task objectives} and \textit{explanation generation}. $g^{\phi}(Tok^{i})$ denotes the probability of $Tok^{i}$ to be an explanation token for the task. $Tok^{1}$ corresponds to the first sub-token of the query sentence and so on. The $[SEP]$ token can never be an explanation therefore it's $GT\equiv 0$. Because of the input length restriction of BERT, here $n=511$}.\label{fig:mtl-nonpool-example}
    \end{figure}

We reiterate that we use the original prediction task as the auxiliary task.
We employ a simple $\mathtt{MLP}$ to map the encoded input $\mathbf{enc}(\mathbf{x})$ to the task prediction $\mathbf{y}$. 
The choice of explanation decoder, however, induces interesting design choices. 
One could in principle pose the generation task as a span detection task or token prediction task.
In this work, we pose the explanation generation as an independent binary classification task over each of the \textit{input words}.
We apply a gated recurrent unit (GRU) over the sequence of output token representations of \bert{} to consider sequential dependencies among tokens.
Then, token representations from the GRU are pooled to form word representations followed by a word-wise $\mathtt{MLP}$. 
Figure~\ref{fig:mtl-nonpool-example} shows the diagram of our proposed approach for the single sentence task (e.g., sentiment detection). 
The same task and explanation generation approach is followed for sentence pair tasks (question-answering) where both the sentences are fed to \bert{}.

    
\subsubsection{Loss function.} 
The explanation loss is composed of individual losses incurred on each input word and can be written as
\begin{equation}
        \mathcal{L}_{exp}=\frac{1}{|S|}\sum_{i=1}^{|S|}|S_{t^i}|\cdot \mathtt{
        BCE}(p^i, t^i)\mbox{,}
\end{equation}
    where $p^i$ is the prediction and $t^i$ is the label of the $i$-th token, $t^i$  equals either to $0$ or to $1$; $|S|$ stands for the length of the passage, $|S_{t^i}|$ is the count of tokens, whose label is the same as $t^i$'s; $\mathtt{BCE}(p, t)$ represents the binary cross-entropy between the prediction $p$ and the label $t$.
    
The overall loss function is the affine combination of the task and explanation prediction. An additional parameter $\lambda$ is used to balance the contribution of both the terms, as shown in the Equation~\ref{eqn:loss_function}.
\begin{equation}
 \label{eqn:loss_function}
 \mathcal{L}_{loss} = \mathcal{L}_{task} + \lambda \mathcal{L}_{exp}\mbox{,}
\end{equation}

where $\mathcal{L}_{loss}$ is the overall loss and $\mathcal{L}_{task}$ and $\mathcal{L}_{exp}$ are loss functions for the \textbf{task} and \textbf{explanation} respectively. 
$\lambda$ regulates the importance of loss function between task and explanation. 

A key challenge in explanation generation is the presence of sparse labels, i.e., the majority of the input words/tokens are not explanations. This leads to training issues due to the label imbalance that the loss function has to account for.
To account for label sparsity, following \citet{chawla2002smote} we up-weight the \textit{log-likelihood} of rationale, while calculating the binary cross entropy~($\mathtt{BCE}$).
The weights are inverse of the prior probabilities of each class within each input passage, i.e., the inverse proportion of non-rationale tokens in the passage. 

\subsection{Prediction Model}

The input to the prediction phase is the extractive explanation as a masked input $\mathbf{x} \otimes g^{\phi}(\mathbf{x})$.
Specifically, we replace each token that is not in the explanation with a wildcard token (period '.' here). This is necessary to maintain the overall structure of the input text.
Note that since we have a pipelined approach, errors in the explanation generation phase might lead to error magnification in the prediction phase. 
Towards this, rather than considering all input instances for training, we limit ourselves to input instances where the auxiliary task prediction is the same as the actual task label, i.e., $f^\psi(\mathbf{x_i}) == y_i$ for a training instance $(\mathbf{x_i}, y_i)$.
We also choose \bert{} as the network $f^{\theta}$/$f^{\tau}$ that aims to predict the true task label. 
The second-stage model is also validated on such masked inputs. 
But we don't rule out any instance according to the auxiliary model prediction during the validation to reflect what happens during test time.

Mathematically, for an instance
$(\mathbf{x}, \mathbf{t}, y)$, the training function of \approach{} works as per equation.~\ref{eqn:mtl_train}.
\begin{equation} \label{eqn:mtl_train}
\begin{aligned}
    & f^\psi(\mathbf{x}) \xrightarrow{} \: \{0,1\} \\
    & g^\phi(\mathbf{x}) \xrightarrow{} \: \{0,1\}^{|S|} \\
    & f^\theta(\mathbf{x} \otimes g^\phi(\mathbf{x})) \xrightarrow{} \: \{0,1\}, & if f^\psi(\mathbf{\mathbf{x}})=y
\end{aligned}
\end{equation}

The inference is also similar but the output of the auxiliary task predictor is not taken under consideration (eqn.~\ref{eqn:mtl_infer}).

\begin{equation} \label{eqn:mtl_infer}
\begin{aligned}
    & f^\psi(\mathbf{x}) \xrightarrow{} \: \{0,1\} \\
    & g^\phi(\mathbf{x}) \xrightarrow{} \: \{0,1\}^{|S|} \\
    & f^\tau(\mathbf{x} \otimes g^\phi(\mathbf{x})) \xrightarrow{} \: \{0,1\}
\end{aligned}
\end{equation}

%% file: setup-avi.tex
\section{Experimental Evaluation}
\label{sec:eval}

We first describe the experimental setup, baselines, and dataset details. In the next section, we elaborate on the experimental results in detail followed by further analysis.

\subsection{Datasets}
We consider three diverse language tasks for our evaluation from the benchmark in~\citet{deyoung2019eraser}. All datasets are split in the same way as provided in the benchmark.
Since we use BERT for representing inputs that have a natural length limitation, we refrain from experimenting with other datasets in the benchmark that contain longer sentences and might require non-trivial input segmentation. Extending our approach to documents with longer sentences and other datasets in the benchmark is left for future work.

\mpara{\movie}~\citet{zaidan2007using,zaidan2008modeling}. One of the original datasets providing extractive rationales, the movies dataset has \textit{pos}itive or \textit{neg}ative sentiment labels on movie reviews. As the included rationale annotations are not necessarily comprehensive (i.e., annotators were not asked to mark \emph{all} text supporting a label), Deyoung et al. collected a comprehensive evaluation set on the final fold of the original dataset~\cite{pang2004sentimental}.
    
\mpara{\fever}~\citet{Thorne18Fever} (short
for Fact Extraction and VERification) is a fact-checking dataset. 
The task is to verify claims from textual sources. 
In particular, each claim is to be classified as \textit{supported}, \textit{refuted} or \textit{not enough information} with reference to a collection of potentially relevant source texts.  
We follow the setup of~\citet{deyoung2019eraser} who restricted this dataset to \textit{supported} or \textit{refuted}.

\mpara{\multirc}~\citet{khashabi2018looking}. This is a reading comprehension dataset composed of questions with multiple correct answers that by construction depend on information from multiple sentences. In \multirc{}, each Rationale is associated with a question, while answers are independent of one another. We convert each rationale/question/answer triplet into an instance within our dataset. Each answer
candidate then has a label of \textit{True} or \textit{False}.

\input{table-hard-sel}

\subsection{Baselines, Competitors, Variants} 

We consider the following competitors that also use a pipelined approach to showcase the effectiveness of our approach
\begin{itemize}
    \item 
    \textbf{\citet{lei2016rationalizing}:}  An end-to-end explain-then-predict approach where rationale generator and decoder are not supervised on rationales;
    
    \item 
    \textbf{\citet{deyoung2019eraser}:} An improvement of the approach of \citet{lei2016rationalizing} where the final loss function has a regularizer based on rationale data. Note that this approach is denoted as Lei et al. (2016) and the previous one is denoted as Lei et al. (2016) (u) in \cite{deyoung2019eraser};
    
    \item 
    \textbf{\citet{lehman2019inferring}:} It is a pipeline approach, where the explanation generation model is trained only on rationales, and the predictor model is trained on ground truth human rationales (instead of on machine predicted rationales as we do) as input to predict the task labels. 
    
    \item 
    \textbf{Bert-To-Bert:} It is implemented in \cite{deyoung2019eraser}, where the generator and the predictor are replaced by \bert{} followed by corresponding MLP heads. It is similar to our \exptask~but we insert an additional GRU layer into the generator, i.e. after the \bert~encoder of the explainer.
    
\end{itemize}

\mpara{Baselines.} In addition to the competitors introduced above, we add two more baselines for better understanding our results -- \cls{} and \human{}.
The \cls{} baseline is trained on the \textit{entire input} to solely optimize for task performance and has no explanation generation functionality. 
The \human{} baseline refers to a prediction model trained just on the \textit{ground-truth human rationales} (all tokens not in the explanation are replaced by a specific wild-card token).

\mpara{\approach{} variants.} Next, we consider three variants of our \approach{} -- (i) \mtlhard{} our original approach, (ii) \exptask{} that only optimizes for explanations in the first stage (does not involve MTL and is task unaware during explanation generation), and (iii) \mtltask{} that reports the auxiliary task performance from the first stage, i.e, it does not involve the second prediction phase.

\subsection{Metrics} 

Mostly denoted as \textit{Perf.} in \cite{deyoung2019eraser}, the \textbf{Macro F1} produced by the \texttt{classification\_score} from sklearn\footnote{https://scikit-learn.org/stable/} is used to evaluate task performance. 
Macro Token-wise F1, presented as \textbf{Token F1} in Table~\ref{tab:hard_task_performance}, is used to measure the proximity of the explanation with human rationales. The precision of an explanation is the fraction of commonly extracted rationale tokens (ER) and ground truth (GT) tokens in comparison to ER. 
While the recall of an explanation is the fraction of common ER tokens with GT in comparison to GT. The Token F1 is the harmonic average of precision and recall of machine rationales.

\subsection{Training setup and Hyper-parameters}
\label{sec:parameter_setup}

All experiments are conducted on an Nvidia 32GB V100 using the PyTorch and Tensorflow framework.
We consider $\bertbase{}$ as the shared encoder model with $\textrm{MAX\_SEQ\_LEN}=512$ and the warm-up proportion $0.1$. 
Both the explanation generation and task prediction models are trained using Adam optimizer~\cite{kingma2014adam} with a batch size of 16, and $\textrm{learning\_rate}=1e-5$. 
Models are trained for 10 epochs with early-stopping criteria on the validation set and $\textrm{patience}=3$.
The MLP for the task classification consists of a dropout layer with a 10\% chance of masking, followed by a $256$ dimensional hidden dense layer, again followed by a Sigmoid output layer. 
The explanation decoder consists of a 128-dimensional GRU with a uniform random kernel analyzer. 
Note that the final outputs of the explanation generator correspond to the sub-token representations of~\bert{}. 
Adjacent sub-tokens are merged to their corresponding original words through max-pooling.
The best $\lambda$ is chosen over a validation set that provides the best trade-off between task performance and token-F1. The best $\lambda$ values for \movie{}, \multirc{}, \fever{} are $5.0, 20.0, 2.0$ respectively.
After training the explanation generation network in \approach{}, we remove instances that the auxiliary output predicts wrongly, and use the rest to train the prediction model. This is to avoid distraction from the wrong predictions from the explanation prediction phase. Note that this is only done during training, while the predictions on the validation and test sets are regardless of the task prediction from the explanation phase.

%% file: table-hard-sel.tex
\begin{table*}[ht!]\begin{minipage}{\linewidth}
    \centering
     
    \begin{tabular}{lcccccc}
        \toprule
        \multirow{2}{*}{\textbf{Approaches}} & \multicolumn{2}{c}{\textbf{\movie}}  & \multicolumn{2}{c}{\textbf{\fever}} & \multicolumn{2}{c}{\textbf{\multirc}}\\
        \cmidrule(lr){2-3}
        \cmidrule(lr){4-5}
        \cmidrule(lr){6-7}
            & \textbf{Macro F1} & \textbf{Token F1} & \textbf{Macro F1} & \textbf{Token F1} & \textbf{Macro F1} & \textbf{Token F1}\\
        \hline 
        \citet{deyoung2019eraser} & 0.914 & 0.285 & 0.719 & 0.234 & 0.655 & 0.456 \\
        \citet{lei2016rationalizing} & \textbf{0.920} & \underline{0.322} & 0.718 & -$^1$ & 0.648 & -$^1$ \\
        \citet{lehman2019inferring} & 0.750 & 0.139 & 0.691 & 0.523 & 0.614 & 0.140 \\
        Bert-To-Bert & 0.860 & 0.145 & 0.877 & \underline{0.812} & 0.633 & 0.412 \\
        
        \hline
        
        \mtltask & 0.884 & \textbf{0.348} & \textbf{0.907} & \textbf{0.837} & \underline{0.718} & \textbf{0.640} \\
        \exptask & 0.814 & 0.142 & 0.795 & 0.801 & \textbf{0.725} & \underline{0.609} \\
        \mtlhard & \underline{0.915} & \textbf{0.348} & \underline{0.894} & \textbf{0.837} & 0.698 & \textbf{0.640} \\
        \hline
        \human & 0.899 & 1.0 & 0.921 & 1.0 & 0.759 & 1.0 \\
        \cls{} & 0.894 & - & 0.916 & - & 0.708 & - \\
        \bottomrule
    \end{tabular}
        
        
    \caption{{Task and explanation performance of hard models, which is defined in section \ref{sec:soft-selection}. Best performances , excluding the \textbf{Token F1} of human annotation, since they are always 1.0, are \textbf{bold} and the second bests are \underline{underlined}. Results for the competitors are kept the same as in the ERASER benchmark~\cite{deyoung2019eraser} whenever it is possible. Also according to \cite{deyoung2019eraser}, $^1$ indicates rationale training degenerated due to the REINFORCE style training.}}
    \label{tab:hard_task_performance}
\end{minipage}
\end{table*}

%% file: results-avi.tex
\section{Results}
\label{sec:results}

We present the results of the effectiveness of our multi-task learning rationale generation framework in Table~\ref{tab:hard_task_performance}. 
Our first observation is that \human{} is quite effective in most datasets and \multirc{} is significantly better than \cls{} in task performance. 
This is perhaps unsurprising because \human{} is trained on extractive rationales that contain task-specific discriminative tokens. 
This also suggests that \cls{} is sometimes distracted by words or tokens unrelated to the task and dropping terms altogether can result in reasonable task performance gains.

Among the variants of \approach{}, \exptask{} model is solely optimized on the explanation loss but has a moderate explanation quality. 
The explanation performance of \approach{} and its variants are the best among all datasets and competitors.
However, it does not perform better than \human{} in terms of task performance. 
This justifies our claim that purely optimizing for explanation accuracy without considering the task context leads to sub-optimal task performance.
Note that \mtltask{} and \mtlhard{} generate the same explanation and have identical explanation quality since they both share the same explanation generation phase.

\exptask{} is outperformed in explanation accuracy (in all datasets) and task accuracy (in \movie{} and \fever{}) by \mtltask{} that jointly optimizes for the task and explanation using shared encoding parameters. 
For \multirc{} and \fever{}, both these variants are already much better than the competitors in task performance but seem less congruent with human rationales.
A crucial difference between our variants with~\citet{lehman2019inferring} and Bert-To-Bert is that the prediction network for those two models is trained over human annotations. However, during the test phase, the output of the machine-generated explanation is considered.
This introduces a distribution mismatch between the training and testing phases.
Unlike them, in both \approach{}~and \exptask~we use the output of the first stage for training the prediction network.

Finally, we present the main result of our paper, i.e., \mtlhard{} and its variants convincingly outperform all other competitors in explanation performance by $\sim 8\%$ on \movie{}, $\sim 5\%$ on \fever{} and more strikingly $\sim 46\%$ on \multirc{}.
Notably, the task performance is at least preserved on~\movie{} or even improved on \fever{} and \multirc{}, compared with other competitors that use joint or rationale data-agnostic training. 
Comparing with~\human{} further verifies our assumption that the models can learn more effectively from rationales data, where the \textit{right reasons} of making predictions are highlighted in advance. 
We attribute this due to two reasons found in our earlier observations -- as in the case of \human{} vs \cls{}, \mtlhard{} being trained on sparser (less noisy) input can predict better. 
Secondly, the explanations are now more contextualized since they are learned along with the task. 
Furthermore, we can see that the task performance can even be sometimes slightly improved by adding a second classifier in the \approach{} compared with the task prediction in the \mtltask (e.g. on \movie{}).


\input{table-soft-sel}

%% file: table-soft-sel.tex
\begin{table}[]
    \centering
    \resizebox{0.98\linewidth}{!}{
    \begin{tabular}{lrcc}
        \toprule
            & Avg. Rationale Len. & Precision & Recall \\
        \hline
        \textbf{\movie}\\
        \citet{deyoung2019eraser} & 8.533   & 0.626 & 0.0333\\
        \citet{lei2016rationalizing} & 430.563 & 0.315 & 0.542\\
        \citet{lehman2019inferring} & 30.530 & 0.505 & 0.102\\
        Bert-To-Bert & 17.500 & 0.614 & 0.072\\
        \exptask & 70.864 & 0.676 & 0.112 \\
        \mtlhard & 86.246 & 0.607 & 0.284 \\
        \human & 240.844 & 1.000 & 1.000 \\
        \hline
        \\
        \textbf{\fever}\\
        \citet{deyoung2019eraser} & 21.894 & 0.438 & 0.35\\
        \citet{lei2016rationalizing} & 138.806 & 0.258 & 0.678\\
        \citet{lehman2019inferring} & 30.882 & 0.584 & 0.508\\
        Bert-To-Bert & 29.127 & 0.904 & 0.811\\
        \exptask & 40.742 & 0.868 & 0.816 \\
        \mtlhard & 44.670 & 0.834 & 0.908 \\
        \human & 39.721 & 1.000 & 1.000\\
        \hline
        \\
        \textbf{\multirc}\\
        \citet{deyoung2019eraser} & 47.699 & 0.337 & 0.352\\
        \citet{lei2016rationalizing} & 155.696 & 0.182 & 0.565\\
        \citet{lehman2019inferring} & 25.150 & 0.245 & 0.118\\
        Bert-To-Bert & 21.699 & 0.726 & 0.326\\
        \exptask & 46.331 & 0.665 & 0.619 \\
        \mtlhard & 55.870 & 0.627 & 0.704 \\
        \human & 49.929 & 1.000 & 1.000 \\
        \bottomrule
    \end{tabular}
    }
    \caption{\small{Statistics of the machine-generated and human-annotated rationales. \textit{Precision} and \textit{Recall} are computed with respect to corresponding human-annotated explanations.}}
    \label{tab:avg_rats_len}
\end{table}


\begin{table}[tb]
    \centering
    \resizebox{\linewidth}{!}{
    \begin{tabular}{lcccc}
    \toprule
        & \textbf{Task} & \textbf{AUPRC} & \textbf{Comp$\uparrow$} & \textbf{Suff$\downarrow$}\\
    \hline
    \textbf{\movie}\\
    BERT-LSTM\\
    + Attention & \textbf{0.970} & 0.417  & 0.129 & \textbf{0.097}\\
    + Gradient & \textbf{0.970} & 0.385 & 0.142 & 0.112\\
    \hline
    \mtlsoft & 0.880 & \textbf{0.420} & \textbf{0.385} & 0.163\\
    \hline
    \textbf{\fever}\\
    GloVe-LSTM\\
    + Attention & 0.870 & 0.235 & 0.037 & 0.122\\
    + Simple Gradient & 0.870 & 0.232 & 0.059 & 0.136\\
    \hline
    \mtlsoft & \textbf{0.914} & \textbf{0.836} & \textbf{0.151} & \textbf{0.068}\\
    \hline
    \textbf{\multirc}\\
    BERT-LSTM\\
    + Attention & 0.655 & 0.244 & 0.036 & 0.052\\
    + Simple Gradient & 0.655 & 0.224 & 0.077 & 0.064\\
    \hline
    \mtlsoft & \textbf{0.726} & \textbf{0.695} & \textbf{0.157} & \textbf{0.031}\\
    \bottomrule
    \end{tabular}
    }
    \caption{\small{Performance of soft models, where the metric of \textit{Task} is \textit{Macro F1}, the same as in Table~\ref{tab:hard_task_performance}, \textit{Comp} represents \textit{Comprehensiveness}, the higher the better and \textit{Suff} is \textit{Sufficiency}, the lower the better.}}
    \label{table:task_explanation_result_soft}
\end{table}

%% file: analysis-avi.tex
\begin{figure*}[tb]
  \resizebox{\textwidth}{!}{
  \subfigure[$\lambda$-selection criteria on \movie]{    \includegraphics[width=0.33\textwidth]{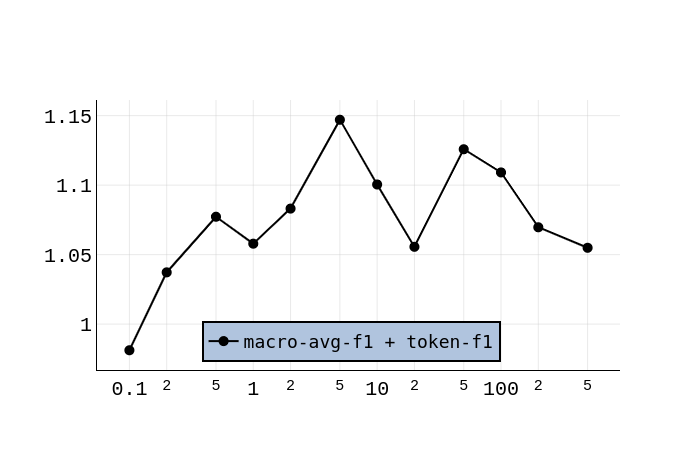}}
  \subfigure[$\lambda$-selection criteria on \fever{}]{    \includegraphics[width=0.33\textwidth]{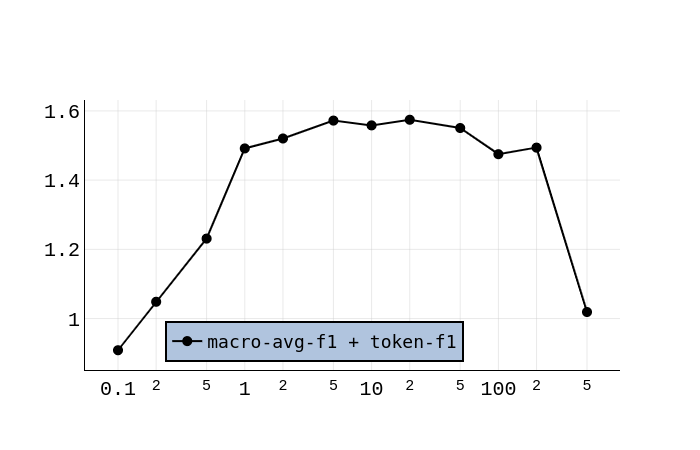}}
  \subfigure[$\lambda$-selection criteria on \multirc]{    \includegraphics[width=0.33\textwidth]{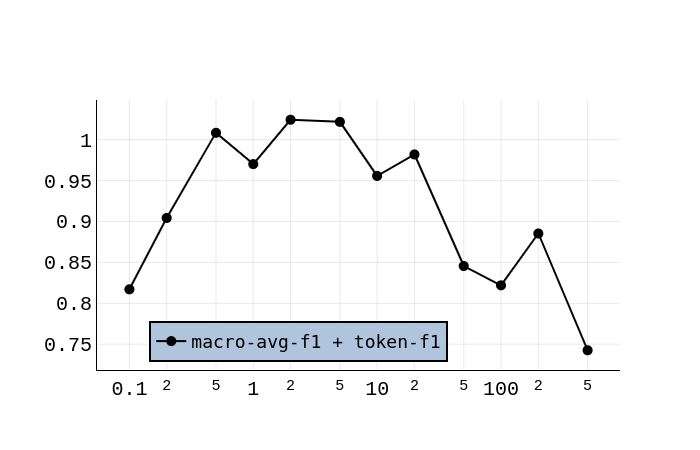}}}
  \caption{$\lambda$ selection criteria for \mtlhard~with $\lambda$ (log-scale) on the validation set of \movie, \fever{} and \multirc. Models for parameter-sweeping is trained on 100\%, 10\% and 25\% of the training set, correspondingly.}
  \label{fig:lambda-sweep-content}
\end{figure*}

\subsection{Effect of $\lambda$}

We have essentially one hyperparameter $\lambda$ from Equation~\ref{eqn:loss_function} that trades-off task and explanation losses during the explanation generation phase.
Since our key objective is to strike an effective balance between task performance and explanation accuracy, we validate our model on a metric that is a simple linear combination of task performance (macro F1) and explanation accuracy (Token F1). We present the effect of $\lambda$ on this combined metric in Figure~\ref{fig:lambda-sweep-content}.

It is evident from the figures that different datasets show different patterns on the metric mixing both task and explanation performance. However, in general, the general trend is that of a steep increase followed by a steep deterioration leave the sweet point balancing the task and explanation performance. 
The $\lambda$ corresponding to the combined metric performance is then selected.

The key takeaway from our experiments on different values of $\lambda$ is that we observe (more-or-less) a stable plateau in the range $\lambda \in [1,50]$ that exhibits low variability performance task and explanation performance.
However, the task performance deteriorates rapidly after $\lambda\geq50$ (or low importance to task-specific loss) indicating that optimizing purely for explanation generation deteriorates task performance.


\subsection{Soft Selection Approaches}
\label{sec:soft-selection}
So far each input word is either a part of an explanation or not. This is categorized as hard-(selection)-model according to \citet{deyoung2019eraser}. It also presents an alternate view to explanations as multi-variable distributions over tokens derived from features, e.g. self-attention values and name it as soft-(selection)-model.
\approach{} can be cast into a soft selection approach explanation model by constructing probability distributions from $g^\phi(\cdot)$ scores of each word before computing the binary cross-entropy.

To evaluate soft selection, the following metrics are used:
\begin{itemize}
    \item 
    \textbf{AUPRC.} or area under the precision-recall curve is used for the soft selection models. Since soft-annotation for each token is assigned with a ranking score (sometimes probability of being rationale).

    \item
     \textbf{Comprehensiveness} of a rationale $r_{ij}$ on instance $i$ and class $j$ is defined as $\mbox{comprehensiveness}(r)=\hat{p}_{ij}-\tilde{p}_{ij}$, where $\hat{p}_{ij}$ is model's prediction on the original input, and $\tilde{p}_{ij}$ is prediction over the input where the rationale $r_{ij}$ is stripped.
     
     \item
     ~\textbf{Sufficiency} on the other hand is defined as the complement of the comprehensiveness, $\mbox{sufficiency}=\hat{p}_{ij}-\bar{p}_{ij}$, where $\bar{p}_{ij}$ is the predicted probability using \emph{only} rationale $r_{ij}$. 

\end{itemize}


Table~\ref{table:task_explanation_result_soft} presents the result of \approach{} in the soft selection mode.
We observe that \methodindv{} performs consistently well both in terms of task and rationale selection metrics. 
A higher value of AUPRC indicates that a better choice of a threshold of per token rationale prediction can help in improving explainability. 
A higher value of comprehensiveness indicates that \methodindv{} selects the correct rationales that are responsible for accurate task label prediction i.e., task performance drops significantly without these tokens. 

The low value of sufficiency also supports the fact i.e., it is an indication that the model can learn the task well only based on those tokens. 
For \movie, BERT-LSTM + Attention can identify rationales well (low sufficiency) and the high value of AUPRC indicates that rationales are following human-annotated ones. 
However, the low value of comprehensiveness reveals that the model can still learn without those rationales. Similar effects were observed in previous work where it was found that attention-based selections are not always rationales~\cite{jain2019attention}.

On the other hand, \methodindv{} ensures that the rationales learned are in accordance with human rationales and the model performance significantly drops without those tokens. 
It fits with our objective that the models should be \textit{interpretable by design}. \methodindv{} performs well both in terms of comprehensiveness and sufficiency for \fever and \multirc. For \movie, \methodindv{} achieves high comprehensiveness but sufficiency is higher (worse) than the baselines. This suggests that \approach{} can retrieve rationale tokens well but those are not sufficient to learn the task, i.e., it fails to capture some rationale tokens. However, it can maintain a balance between task and rationale selection.



\begin{figure*}[ht!]
    \centering
    \includegraphics[width=2.1\columnwidth]{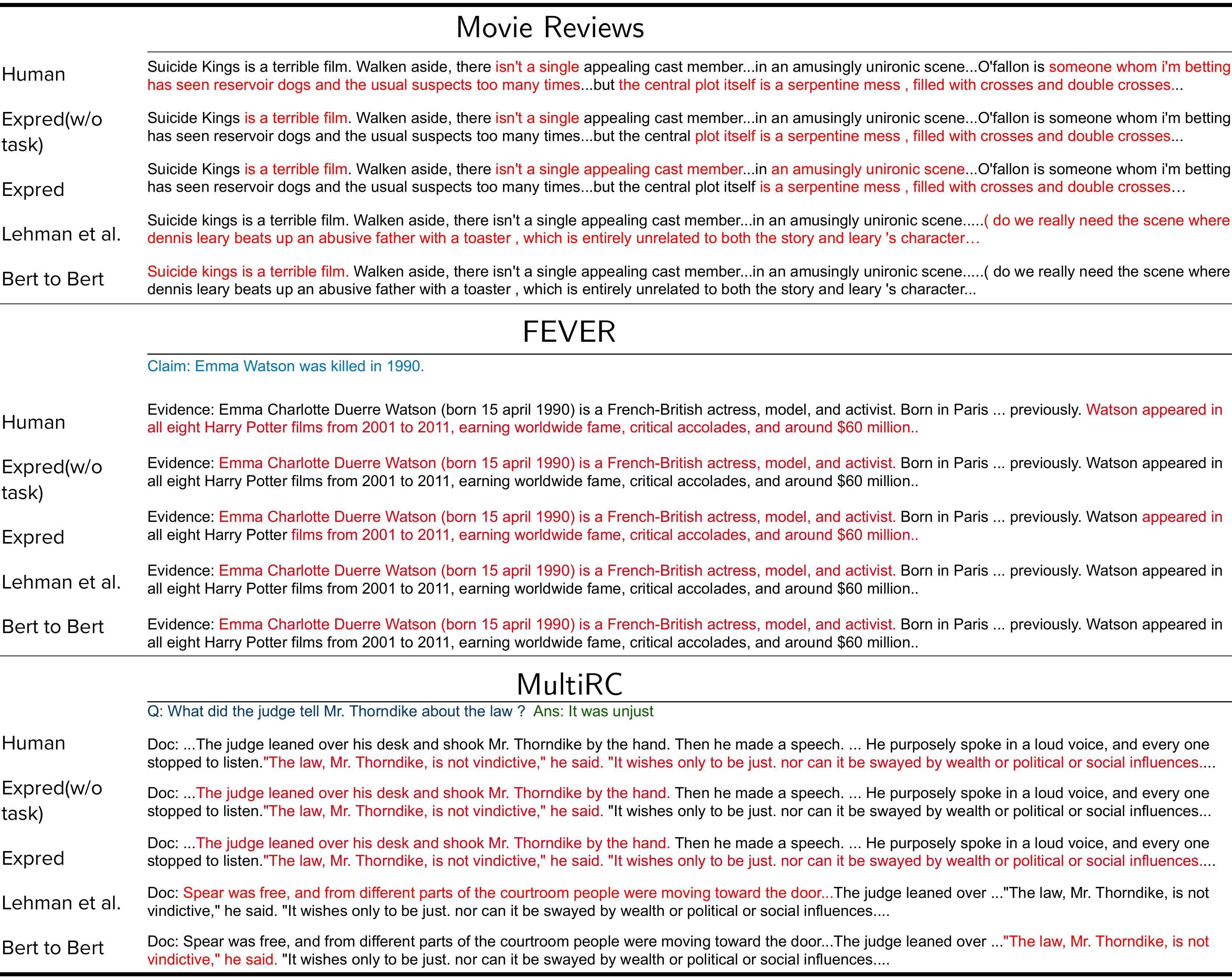}
    \caption{ Anecdotal examples of predictions and explanations by different baselines. Extractive explanations are marked in RED.}\label{fig:anecdotal_examples}
    \vspace{-1mm}
\end{figure*}

\subsection{Machine explanations vs Human explanation}

From the previous results, it is tempting to conclude that we improve task performance at the expense of being less congruent with human rationales and vice versa. 
Towards getting a clearer understanding we perform some further analysis to compare explanations generated by our approach vs human rationales. 
We present the results of our analysis in anecdotal example in Figure~\ref{fig:anecdotal_examples} and explanation statistics in Table~\ref{tab:avg_rats_len}. 
First, we observe that for \movie{} our generated explanations are far shorter (avg. length of 86.246 words) in length than those annotated by humans (avg. length of 240 words). 
For this dataset, we also observe that while \approach{} generates explanations that are \textit{sufficiently} predictive, human explanations tend to be more comprehensive. 
This is also supported by the relatively high precision and low recall.
From the anecdotal evidence, we see evidence of this fact where human annotations are far more verbose than any of the baselines.
Unlike \movie{}, the precision and the recall of the \approach{} explanations are the most balanced for the other datasets compared to other baseline models. This in turn results in higher F1 values as presented in the Table \ref{tab:hard_task_performance}.

Comparing the explanations from other baselines, we observe that \mtlhard{} tends to be more comprehensive (yet sparse) than Lehmann et al.~\cite{lehman2019inferring} and its Bert variant Bert-to-Bert.
This suggests that the sparsity constraints in Lehman et al~\cite{lehman2019inferring} prevent the model from learning comprehensive explanations and also have an effect on task performance. 
We on the other hand do not have explicit regularizers on sparsity.

Finally, as an artifact of the human annotation process, we see that the explanations collected can sometimes be noisy due to the under-specified and ambiguous nature of the task definition. 
Specifically, for \movie{} we observe some predictive phrases are missed by humans, and other phrases that do not contribute substantial predictive value are annotated.
However, these rationales, though noisy, still hold a lot of value for learning better models as is exemplified by our results.
Moreover, the lower Token-F1 score should not be misconstrued with a lack of interpretability rather than deviations from human rationales.
Due to this comprehensiveness and sufficiency between human and machine explanations~\cite{strout2019human} proposes a further human evaluation of the machine-generated explanations. 
Since our objective in this paper is to generate proper rationales that are sufficient to make predictions, such human evaluation is left for future work.

%% file: 07.conclusions.tex
\section{Conclusions}
\label{conclusion}

In this paper we propose a novel yet simple approach \approach{}, that uses multi-task learning in the explanation generation phase to provide better task-aware explanations for explain-then-predict models.
We find that we substantially outperform existing explain-then-predict approaches by 7\% - 47\% by explicitly incorporating task-specific supervision during explanation generation. Additionally, we observed that we can also use \approach{} in the soft selection setting and observe competitive results.
Our main observation is that simple pipeline models like \mtlhard{} can indeed strike a good balance between explanation quality and task performance, consistently performing at par or even better than models when given full inputs.
This is in contrast to joint models like~\cite{lei2016rationalizing} that find it hard to incorporate rationales data and are hard to train in general and difficult to maintain.

There are many avenues for future work that are possible. First, end-to-end models outperform \mtlhard{} in task performance for~\movie{} dataset indicating that for some tasks rationale data might be limited or might not be sufficient to deliver better task performance.
We would want to scale rationale collection methods and study the impact of the size of the rationale dataset on task performance. We would also want to extend our current pipelined approach to an end-to-end approach. 
Finally, an important open question that this work prompts is that can extractive explanations be generalized to other Web tasks like search~\cite{singh2016expedition, holzmann2017exploring} and structured data~\cite{fetahu2019tablenet}.